\documentclass[conference]{IEEEtran}
\IEEEoverridecommandlockouts

\usepackage{cite}
\usepackage{amsmath,amssymb,amsfonts}
\usepackage{algorithmic}
\usepackage{graphicx}
\usepackage{textcomp}
\usepackage{xcolor}
\usepackage{bm}
\usepackage{booktabs} 
\usepackage{siunitx} 
\usepackage{amssymb}  
\usepackage{array}
\usepackage{multirow}
\newcolumntype{C}{>{\centering\arraybackslash}p{1.2cm}}
\newcolumntype{A}{>{\centering\arraybackslash}p{1.53cm}}
\newcolumntype{B}{>{\centering\arraybackslash}p{1.72cm}}
\def\BibTeX{{\rm B\kern-.05em{\sc i\kern-.025em b}\kern-.08em
    T\kern-.1667em\lower.7ex\hbox{E}\kern-.125emX}}
\makeatletter
\newcommand{\linebreakand}{%
  \end{@IEEEauthorhalign}
  \hfill\mbox{}\par
  \mbox{}\hfill\begin{@IEEEauthorhalign}
}
\makeatother
    
\begin{document}


\title{Diff-KD: Diffusion-based Knowledge Distillation for Collaborative Perception under Corruptions
\thanks{*Corresponding author: Chaokun Zhang (zhangchaokun@tju.edu.cn).}
} 

\author{\IEEEauthorblockN{1\textsuperscript{st} Pengcheng Lyu}
\IEEEauthorblockA{\textit{School of Future Technology} \\
\textit{Tianjin University}\\
Tianjin, China \\
lvpc\_0726@tju.edu.cn}
\and
\IEEEauthorblockN{2\textsuperscript{nd} Chaokun Zhang*}
\IEEEauthorblockA{\textit{School of Cybersecurity} \\
\textit{Tianjin University}\\
Tianjin, China \\
zhangchaokun@tju.edu.cn}
\linebreakand  
\IEEEauthorblockN{3\textsuperscript{rd} Gong Chen}
\IEEEauthorblockA{\textit{School of Computer Science and Technology} \\
\textit{Tianjin University}\\
Tianjin, China \\
gongchen01@tju.edu.cn}
\and
\IEEEauthorblockN{4\textsuperscript{th} Tao Tang}
\IEEEauthorblockA{\textit{School of Computer Science and Technology} \\
\textit{Tianjin University}\\
Tianjin, China \\
tangtao@tju.edu.cn}
\and
\IEEEauthorblockN{5\textsuperscript{th} Zhaoxiang Luo}
\IEEEauthorblockA{\textit{School of Future Technology} \\
\textit{Tianjin University}\\
Tianjin, China \\
lzx0618@tju.edu.cn}
}

\maketitle

\begin{abstract} 
Multi-agent collaborative perception enables autonomous systems to overcome individual sensing limits through collective intelligence. However, real-world sensor and communication corruptions severely undermine this advantage. Crucially, existing approaches treat corruptions as static perturbations or passively conform to corrupted inputs, failing to actively recover the underlying clean semantics. To address this limitation, we introduce Diff-KD, a framework that integrates diffusion-based generative refinement into teacher-student knowledge distillation for robust collaborative perception. Diff-KD features two core components: (i) Progressive Knowledge Distillation (PKD), which treats local feature restoration as a conditional diffusion process to recover global semantics from corrupted observations; and (ii) Adaptive Gated Fusion (AGF), which dynamically weights neighbors based on ego reliability during fusion. Evaluated on OPV2V and DAIR-V2X under seven corruption types, Diff-KD achieves state-of-the-art performance in both detection accuracy and calibration robustness.
\end{abstract}

\begin{IEEEkeywords} 
Collaborative perception, knowledge distillation, diffusion models, robustness to corruptions
\end{IEEEkeywords}

\section{Introduction} 
\label{sec:intro}

Multi-agent collaborative perception has emerged as a cornerstone technology for next-generation autonomous systems, enabling vehicles or robots to transcend the limitations of individual sensors by fusing observations from distributed agents~\cite{bevformer}. This collective intelligence dramatically expands perceptual coverage and enhances robustness, enabling safer autonomy in complex environments~\cite{review}.
However, translating this theoretical advantage into reliable real-world performance remains a formidable challenge. In practical deployment, agents often operate under corrupted local observations~\cite{RCP-Bench,v2x-dgw,rocooper}, which may result from adverse weather conditions (such as fog or heavy rain), internal sensor failures (such as LiDAR beam missing or motion blur), or external disturbances (such as electromagnetic interference or echo interference). When such imperfect data is shared and fused across the network, it can introduce erroneous context or even amplify uncertainty, ultimately degrading the system’s overall accuracy and safety.

\begin{figure*}[t]
  \centering
  \includegraphics[width=1.0\linewidth, clip, trim=0cm 2.2cm 0cm 2.2cm]{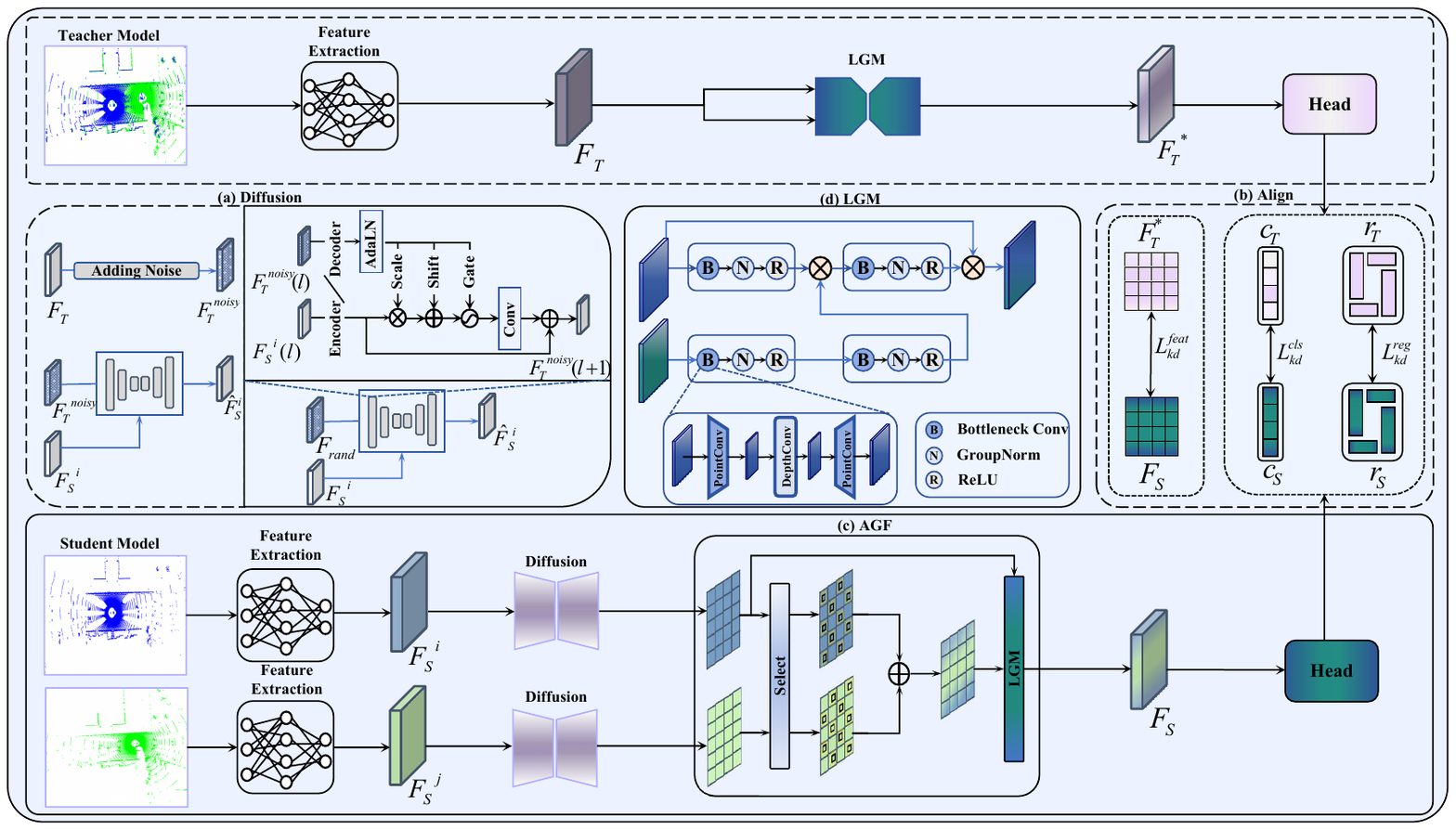}
  \caption{Overall architecture of Diff-KD. The framework comprises a teacher model (input: holistic-view point cloud) and a student model (input: multi-agent local point clouds). Knowledge transfer is achieved via Progressive Knowledge Distillation (PKD), including: \textbf{(a)} active feature restoration using diffusion before fusion; and \textbf{(b)} alignment of features and predictions after fusion. Furthermore, the student employs an \textbf{(c)} Adaptive Gated Fusion (AGF) module that fuses features from multiple agents, incorporating a \textbf{(d)} Lightweight Gated Modulation block (LGM) — the same LGM architecture is also applied in the teacher model to enhance its global features. Dashed components are active only during training.}
  \label{fig:framework}
\end{figure*}

Existing approaches to mitigate these issues typically fall into two broad categories. The first category relies on robust fusion architectures that attempt to suppress or discard noisy inputs during aggregation. V2X-ViT~\cite{v2x-vit} leverages heterogeneous multi-agent attention to adaptively fuse features across vehicles and infrastructure. ERMVP~\cite{ermvp} suppresses pose-induced misalignment and noise via sparse consensus calibration. However, these fusion-centric methods treat local observations as given and lack mechanisms to actively recover missing semantics or reconstruct corrupted features.
The second category adopts teacher-student knowledge distillation, wherein a powerful centralized teacher model guides the training of lightweight student model. DiscoNet~\cite{disconet} distills knowledge from an early-fusion teacher to train an intermediate-fusion student. DSRC~\cite{dsrc} introduces a sparse-to-dense distillation framework to learn density-insensitive and semantic-aware representations against real-world corruptions. However,  such approaches passively mimic the teacher’s outputs or intermediate features, treating corrupted observations as immutable inputs and failing to actively restore the underlying clean semantics. 

To address these limitations, we propose Diff-KD, a novel teacher-student framework that \textit{actively} transforms noisy, incomplete local observations into globally consistent representations through two synergistic mechanisms.
The first is Progressive Knowledge Distillation (PKD). We treat feature refinement as a generative task. Before fusion, each agent’s corrupted local features are independently restored toward the teacher’s global view using a conditional diffusion-based denoiser~\cite{coopdiff,conditional}, conditioned on its own observation. After fusion, both the student’s fused features and final predictions are aligned with their corresponding teacher counterparts, ensuring semantic consistency across the entire distillation pipeline.
The second is Adaptive Gated Fusion (AGF), which dynamically balances ego-centric reliability and neighbor-supplied complementarity via multi-agent adaptive weighting and a lightweight gated modulation block (LGM).

To rigorously evaluate robustness under realistic sensing and communication imperfections, we adopt the physically grounded corruption models from recent 3D robustness benchmarks~\cite{robo3d} and apply them to the original OPV2V~\cite{opv2v} and DAIR-V2X~\cite{dair-v2x} datasets, which systematically simulate the aforementioned corruption types.
Crucially, Diff-KD is trained exclusively on clean data and evaluated on all corruption variants without any fine-tuning. Despite this challenging setting, our method consistently outperforms all existing state-of-the-art approaches across every corruption type, demonstrating its exceptional generalization capability and resilience to input corruptions.

The main contributions can be summarized as follows:
\begin{itemize}
    \item We propose Diff-KD, a novel teacher–student collaborative perception framework that leverages generative  refinement to actively restore corrupted local features toward a globally consistent teacher representation.

    \item We introduce two synergistic modules: (i) PKD, which performs pre-fusion feature denoising via a conditional diffusion process and post-fusion alignment of both features and predictions; and (ii) AGF, which dynamically balances ego reliability and neighbor complementarity under uncertainty.

    \item Extensive experiments on OPV2V and DAIR-V2X under diverse physically grounded corruptions demonstrate that Diff-KD achieves state-of-the-art accuracy and robustness.
\end{itemize}

\section{Method}

As illustrated in Fig.~\ref{fig:framework}, our framework adopts a teacher–student knowledge distillation paradigm to guide the student model toward learning a globally consistent and domain-invariant feature representation.

Considering a collaborative scene with $N$ agents, the teacher model first performs spatial alignment and fusion of point clouds from all agents to construct a unified global-view point cloud.
This fused point cloud is voxelized and processed by a PointPillars LiDAR encoder~\cite{pointpillars} to produce a Bird’s Eye View (BEV) feature map, denoted as $\bm{F}_{\mathrm{T}}$.
To further enhance its representational capacity, we apply the Lite Gated Modulation (LGM), yielding an augmented teacher feature $\bm{F}_{\mathrm{T}}^{*}$, which is subsequently fed into the detection head to generate classification and regression outputs $(\bm{c}_{\mathrm{T}}, \bm{r}_{\mathrm{T}})$.

The student model shares most of its design with the teacher model, e.g., the backbone encoder, the LGM block, and the detection head. 
Each agent $i$ independently extracts its local BEV feature $\bm{F}_{\mathrm{S}}^{i}$.  
These features are then refined via a conditional diffusion process that is trained to reconstruct the teacher’s global feature $\bm{F}_{\mathrm{T}}$, performing denoising and semantic completion to yield the refined local features $\{\hat{\bm{F}}_{\mathrm{S}}^{i}\}_{i=1}^{N}$. 
These refined features are subsequently aggregated by a feature fusion module (described in Section~\ref{subsec:agf}) to produce the fused student feature $\bm{F}_{\mathrm{S}}$. This fused representation is aligned with $\bm{F}_{\mathrm{T}}^{*}$. 
Finally, $\bm{F}_{\mathrm{S}}$ is fed into the detection head to yield predictions $(\bm{c}_{\mathrm{S}}, \bm{r}_{\mathrm{S}})$,  which are aligned with $(\bm{c}_{\mathrm{T}}, \bm{r}_{\mathrm{T}})$.  
Upon completion of the full distillation process, only the student model is deployed for inference.

\subsection{Progressive Knowledge Distillation}

We formalize the distillation process as Progressive Knowledge Distillation (PKD), which consists of diffusion-based local feature refinement before fusion to recover corrupted semantics at the agent level and alignment of both intermediate features and final predictions between the student and teacher models after fusion to ensure global semantic consistency between the student and teacher models.

\noindent
\textbf{Diffusion-Based Local Feature Refinement.}
To enhance the student model's generalization capability under adverse conditions, we formulate inter-agent knowledge transfer as a conditional denoising generative task. Each agent’s local representation is independently refined prior to fusion.
Specifically, the teacher's globally consistent BEV feature $\bm{F}_{\mathrm{T}}$ is treated as the “clean ground-truth” representation, while each student agent's local feature $\bm{F}_{\mathrm{S}}^{i}$ serves as a conditional signal that guides a diffusion model to recover a teacher-like feature from noise.

During training, we first apply the standard forward diffusion process to $\bm{F}_{\mathrm{T}}$. Given a total of $\mathrm{T}$ diffusion steps, at a randomly sampled time step $t \sim \mathcal{U}(1, T)$, Gaussian noise is injected as: 
\begin{equation}
\bm{F}_{\mathrm{T}}^{\mathrm{noisy}} = \sqrt{\bar{\alpha}_t} \, \bm{F}_{\mathrm{T}} + \sqrt{1 - \bar{\alpha}_t} \, \bm{\epsilon}, \quad \bm{\epsilon} \sim \mathcal{N}(0, \mathbf{I}),
\label{eq:diffusion_forward}
\end{equation}
where $\bar{\alpha}_{t}$ is the noise scaling factor at time step $t$.
The noisy teacher feature $\bm{F}_{\mathrm{T}}^{\mathrm{noisy}}$ is then denoised in a conditioned diffusion model, where the student’s local feature $\bm{F}_{\mathrm{S}}^{i}$ acts as the conditioning input. Specifically, we employ a conditional BiFPN-like~\cite{BiFPN} diffusion architecture, in which the conditioning signal is injected via Conditional Adaptive Modulation (CAM), a FiLM-like operation~\cite{film,diffuser} adapted to diffusion models. In CAM, the condition $\bm{F}_{\mathrm{S}}^{i}$ is processed through an AdaLN pathway to dynamically generate modulation parameters—Scale, Shift, and Gate—which are applied to the noisy feature at each layer:
\begin{equation}
\bm{F}_{\mathrm{T}}^{\mathrm{noisy}}(l+1),\, \bm{F}_{\mathrm{S}}^{i}(l+1) = \mathrm{CAM}\left( \bm{F}_{\mathrm{T}}^{\mathrm{noisy}}(l),\, \bm{F}_{\mathrm{S}}^{i}(l) \right),
\label{eq:cam_update}
\end{equation}
where $l$ denotes the layer index in the multi-level feature extraction hierarchy. After a complete forward pass through the diffusion network, the refined feature $\hat{\bm{F}}_{\mathrm{S}}^{i}$ is obtained by iteratively denoising the input according to the DDIM~\cite{ddim} sampling scheme.
The diffusion loss is defined as the mean squared error between the predicted and true noise:
\begin{equation}\mathcal{L}_{\mathrm{diff}}^{i}=\mathbb{E}_{t,\bm{\epsilon}}\left[\left\|\bm{\epsilon}-\hat{\bm{\epsilon}}_\theta\left(\bm{F}_\mathrm{T}^\mathrm{noisy},\bm{F}_{\mathrm{S}}^{i},t\right)\right\|_2^2\right],
\end{equation}
where $\hat{\epsilon}_\theta$ denotes the noise prediction network.

At inference time, the teacher model is no longer available. The diffusion module thus functions as a generative feature optimizer: starting from pure Gaussian noise $\bm{F}_{\mathrm{rand}}\sim\mathcal{N}(0,\mathbf{I})$, it leverages only the student’s own local feature $\bm{F}_{\mathrm{S}}^{i}$ as the conditioning signal to synthesize an enhanced representation $\hat{\bm{F}}_{\mathrm{S}}^{i}$.

\noindent
\textbf{Feature and Prediction Alignment.}
After refinement, the set of enhanced local features $\{ \hat{\bm{F}}_{\mathrm{S}}^{i} \}_{i=1}^{N}$ is fused into a global student representation $\bm{F}_{\mathrm{S}}$.  To ensure holistic consistency with the teacher, we impose two levels of knowledge distillation: feature-level and output-level.
The feature-level alignment is computed between $\bm{F}_{\mathrm{T}}^{*}$ and $\bm{F}_{\mathrm{S}}$ and the distillation loss is defined via the Kullback–Leibler divergence:
\begin{equation}
    \mathcal{L}_{\mathrm{kd}}^{\mathrm{feat}} = \mathrm{KL}\big( \mathrm{softmax}(\bm{F}_{\mathrm{S}}) \,\|\, \mathrm{softmax}(\bm{F}_{\mathrm{T}}^{*}) \big),
    \label{eq:kd_feat}
\end{equation}
where softmax is applied channel-wise across the feature dimension.
Similarly, we apply KL-based alignment to $(\bm{c}_{\mathrm{T}}, \bm{r}_{\mathrm{T}})$ and $(\bm{c}_{\mathrm{S}}, \bm{r}_{\mathrm{S}})$, yielding $\mathcal{L}_{\mathrm{kd}}^{\mathrm{cls}}$ and $\mathcal{L}_{\mathrm{kd}}^{\mathrm{reg}}$, respectively.
The post-fusion knowledge distillation loss is then formed by combining these three components:
\begin{equation}
    \mathcal{L}_{\mathrm{kd}}^{\mathrm{post}} = \mathcal{L}_{\mathrm{kd}}^{\mathrm{feat}} + \mathcal{L}_{\mathrm{kd}}^{\mathrm{cls}} + \mathcal{L}_{\mathrm{kd}}^{\mathrm{reg}},
    \label{eq:kd_post}
\end{equation}
and the overall PKD objective integrates both pre-fusion and post-fusion stages:
\begin{equation}\mathcal{L}_{\mathrm{PKD}}=\sum_{i=1}^N\mathcal{L}_{\mathrm{diff}}^{i}+\mathcal{L}_{\mathrm{kd}}^{\mathrm{post}}.\end{equation}

Notably, the total training objective further incorporates standard detection losses on the student’s predictions to ensure task fidelity: a classification loss $\mathcal{L}_{\mathrm{cls}}$ computed via sigmoid focal loss, and a regression loss $\mathcal{L}_{\mathrm{reg}}$ implemented as weighted smooth L1 loss. Together with $\mathcal{L}_{\mathrm{PKD}}$, they form the final objective
\begin{equation}
    \mathcal{L}_{\mathrm{final}} = \mathcal{L}_{\mathrm{PKD}} + \mathcal{L}_{\mathrm{cls}} + \mathcal{L}_{\mathrm{reg}}.
\end{equation}

\begin{table*}[!t]
\centering
\caption{Overall performance on OPV2V and DAIR-V2X under clean and corrupted conditions (AP@0.5 / AP@0.7).}
\label{tab:overall performance}
\begin{tabular}{l|A B A A A B A A}
\toprule
\textbf{Corruptions} & \textbf{Clean} & \textbf{Beam Missing} & \textbf{Motion Blur} & \textbf{Fog} & \textbf{Cross Talk} & \textbf{Cross Sensor} & \textbf{Water} & \textbf{Echo} \\
\midrule
\multicolumn{9}{c}{\textit{\textbf{OPV2V}}} \\
\midrule
No Collaboration & 73.81 / 50.84 & 66.64 / 41.57 & 60.15 / 27.77 & 53.76 / 40.08 & 63.75 / 38.48 & 60.46 / 36.88 & 72.09 / 46.18 & 66.98 / 46.04 \\
Late Fusion~\cite{latefusion}      & 86.69 / 80.24 & 84.00 / 74.49 & 77.65 / 53.60 & 70.34 / 63.38 & 83.07 / 73.87 & 80.06 / 68.54 & 86.91 / 79.88 & 82.71 / 76.38 \\
ERMVP~\cite{ermvp}            & 91.37 / 84.01 & 86.47 / 76.88 & 85.26 / 66.15 & 66.66 / 57.29 & 81.59 / 71.04 & 80.49 / 69.81 & 89.97 / 81.02 & 89.51 / 82.18 \\
V2X-ViT~\cite{v2x-vit}          & 90.38 / 81.36 & 86.75 / 74.95 & 74.53 / 53.98 & 63.52 / 55.67 & 77.22 / 65.75 & 80.80 / 67.12 & 88.51 / 78.45 & 89.24 / 80.08 \\
Fcooper~\cite{fcooper}          & 90.42 / 82.50 & 86.38 / 75.36 & 73.04 / 49.96 & 66.87 / 59.64 & 78.03 / 65.61 & 80.03 / 67.29 & 89.34 / 80.47 & 87.90 / 80.00 \\
Where2Comm\cite{where2comm}       & 89.63 / 79.06 & 84.27 / 70.89 & 82.15 / 58.54 & 69.66 / 60.05 & 87.00 / 73.99 & 77.56 / 63.54 & 88.28 / 75.72 & 88.55 / 77.43 \\
DSRC\cite{dsrc}             & 91.82 / 85.23 & 88.04 / 79.13 & 84.87 / 64.60 & 71.01 / 64.23 & 80.43 / 70.34 & 81.91 / 71.56 & 89.43 / 84.70 & 90.85 / 84.12 \\
\textbf{Diff-KD (Ours)}    & \textbf{92.03 / 87.81} & \textbf{87.86 / 82.27} & \textbf{86.17 / 70.57} & \textbf{71.04 / 64.57} & \textbf{87.71 / 81.27} & \textbf{81.94 / 75.91} & \textbf{90.01 / 85.24} & \textbf{91.72 / 87.67} \\
\midrule
\multicolumn{9}{c}{\textit{\textbf{DAIR-V2X}}} \\
\midrule
No Collaboration & 66.13 / 51.80 & 33.70 / 25.98 & 58.83 / 37.13 & 40.79 / 30.57 & 59.82 / 41.64 & 34.59 / 26.29 & 59.98 / 49.79 & 63.29 / 50.66 \\
Late Fusion~\cite{latefusion}      & 65.25 / 55.79 & 33.09 / 27.96 & 53.66 / 38.52 & 41.49 / 35.74 & 59.22 / 48.72 & 33.39 / 27.75 & 57.88 / 40.89 & 61.11 / 43.87 \\
ERMVP~\cite{ermvp}            & 67.59 / 55.65 & 34.20 / 27.79 & 55.81 / 39.60 & 41.82 / 34.31 & 59.29 / 45.96 & 34.94 / 27.94 & 62.12 / 49.22 & 66.78 / 54.61 \\
V2X-ViT~\cite{v2x-vit}          & 71.28 / 54.04 & 37.79 / 26.56 & 62.14 / 39.32 & 44.11 / 33.29 & 65.24 / 45.15 & 36.65 / 26.26 & 64.28 / 46.48 & 69.19 / 51.90 \\
Fcooper~\cite{fcooper}          & 73.97 / 56.14 & 40.29 / 28.80 & 60.66 / 35.20 & 43.52 / 32.18 & 66.36 / 45.71 & 38.00 / 27.69 & 67.29 / 49.18 & 72.91 / 55.41 \\
Where2Comm~\cite{where2comm}       & 67.41 / 53.15 & 35.95 / 28.03 & 55.62 / 36.77 & 42.00 / 33.40 & 61.35 / 44.61 & 35.41 / 27.03 & 62.10 / 46.29 & 66.60 / 51.48 \\
DSRC~\cite{dsrc}             & 75.21 / 61.62 & 43.50 / 32.73 & 64.50 / 44.55 & 46.72 / 37.30 & 64.89 / 50.17 & 42.13 / 30.84 & 69.21 / 53.58 & 74.01 / 60.60 \\
\textbf{Diff-KD (Ours)}    & \textbf{78.27 / 63.92} & \textbf{48.15 / 33.05} & \textbf{70.21 / 49.02} & \textbf{48.53 / 38.28} & \textbf{71.70 / 53.75} & \textbf{43.00 / 31.96} & \textbf{70.48 / 54.51} & \textbf{77.11 / 62.90} \\
\bottomrule
\end{tabular}
\end{table*}

\subsection{Adaptive Gated Fusion}
\label{subsec:agf}

To effectively integrate multi-view information while suppressing potential artifacts introduced by the diffusion process, we propose the Adaptive Gated Fusion (AGF) module. AGF performs pixel-wise adaptive weighting to aggregate features from all agents, preserving the dominance of the ego-agent’s representation while modulating collaborative information through learnable gating signals.

Given the refined feature set $\{\hat{\bm{F}}_{\mathrm{S}}^{i} \in \mathbb{R}^{C \times H \times W}\}_{i=1}^{N}$ with $\hat{\bm{F}}_{\mathrm{S}}^{1}$ as the ego feature, we compute agent-wise importance maps via a lightweight convolutional network $\mathcal{P}(\cdot)$:
\begin{equation}
    \alpha_i = \mathcal{P}\left( \left[ \hat{\bm{F}}_{\mathrm{S}}^{i} \,;\, \hat{\bm{F}}_{\mathrm{S}}^{1} \right] \right) \in \mathbb{R}^{1 \times H \times W}, \quad i = 1, \dots, N.
    \label{eq:importance_map}
\end{equation}

Adaptive weights are computed per spatial location by applying softmax across the agent dimension:
\begin{equation}
\omega_{i}(p) = \frac{\exp\left(\alpha_{i}(p)\right)}{\sum_{j=1}^{N} \exp\left(\alpha_{j}(p)\right)}, \quad \forall p \in \{1, \dots, H \times W\}.
\label{eq:softmax_weights}
\end{equation}

The channel-broadcasted version of $\omega_i$ is denoted as $\Omega_i$, and the spatially adaptive collaborative feature is then computed as a weighted fusion via element-wise multiplication:
\begin{equation}
\bm{F}_{\mathrm{optimal}} = \sum_{i=1}^{N} \Omega_{i} \odot \hat{\bm{F}}_{\mathrm{S}}^{i},
\label{eq:f_optimal}
\end{equation}

To further enable fine-grained fusion between the ego feature and the collaborative consensus, we employ a lightweight gated modulation block\cite{scsegamba} (structurally identical to the teacher's LGM). This block takes $\hat{\bm{F}}_{\mathrm{S}}^{1}$ as the trunk and $\bm{F}_{\mathrm{optimal}}$ as the condition. It employs multi-stage bottleneck convolutions ($\mathcal{B}_{i}(\cdot)$) and group normalization layers ($\mathcal{N}_{i}(\cdot)$) to construct a spatial-channel joint gating mechanism:
\begin{equation}
\bm{C}_{1} = \mathrm{ReLU}\left( \mathcal{N}_{1}\left( \mathcal{B}_{1}( \bm{F}_{\mathrm{optimal}} ) \right) \right),
\end{equation}
\begin{equation}
\bm{C}_{2} = \mathrm{ReLU}\left( \mathcal{N}_{2}\left( \mathcal{B}_{2}( \bm{C}_{1} ) \right) \right).
\end{equation}
Each $\mathcal{B}_{i}(\cdot)$ consists of a 1×1 convolution (for channel reduction), a 3×3 depthwise convolution (for spatial modeling), and another 1×1 convolution (for channel restoration), approximating a low-rank decomposition to significantly reduce computational complexity. The output $\bm{C}_{2}$ serves as the gating coefficient, dynamically reflecting the reliability of collaborative information at each spatial location. Concurrently, the ego feature is transformed into a modulatable base representation:
\begin{equation}
\hat{\bm{X}} = \mathrm{ReLU}\left( \mathcal{N}_{3}\left( \mathcal{B}_{3}( \hat{\bm{F}}_{\mathrm{S}}^{1} ) \right) \right),
\end{equation}
which is then element-wise multiplied with the gating coefficient:
\begin{equation}
\hat{\bm{X}}_{\mathrm{fused}} = \hat{\bm{X}} \odot \bm{C}_{2}.
\end{equation}

 Finally, the fused feature is refined by an additional Bottleneck Convolution to enhance representational capacity and combined with a residual connection:
\begin{equation}
\bm{F}_{\mathrm{S}} = \mathrm{ReLU}\left( \mathcal{N}_{4}\left( \mathcal{B}_{4}( \hat{\bm{X}}_{\mathrm{fused}} ) \right) \right) + \hat{\bm{F}}_{\mathrm{S}}^{1}.
\end{equation}

\section{Experiments}

\subsection{Datasets and Experimental Settings}

\noindent
\textbf{Datasets.}
We evaluate our method on two large-scale collaborative perception benchmarks: OPV2V~\cite{opv2v}, a simulated dataset with diverse urban and highway scenarios, and DAIR-V2X~\cite{dair-v2x}, a real-world vehicle-to-infrastructure dataset. To assess robustness under realistic sensing and communication corruptions, we adopt the seven physically grounded corruption types from Robo3D~\cite{robo3d} and apply them to both datasets. These corruptions include beam missing, motion blur, fog, cross talk, cross sensor, water, and echo.

\noindent
\textbf{Experimental Settings.}
We implement the proposed and comparative models using the OpenCOOD framework~\cite{opv2v} and train them on a single NVIDIA RTX 4090 (24 GB) GPU with Adam (initial learning rate $10^{-3}$). All models are trained for 40 epochs with a batch size of 2 on the clean training splits. We set the number of collaborative vehicles to 2 for DAIR-V2X and 5 for OPV2V. In our method, both the diffusion encoder and decoder consist of three layers, using 10 denoising sampling steps.

\subsection{Quantitative Evaluation}

\noindent
\textbf{Detection Performance.}
Table~\ref{tab:overall performance} presents the 3D object detection performance of our method on OPV2V and DAIR-V2X, compared against a non-collaborative baseline and several collaborative approaches. The baseline, No Collaboration, uses only the ego vehicle's data for detection. Collaborative methods include Late Fusion~\cite{latefusion} (which aggregates per-agent detection outputs) and state-of-the-art intermediate fusion frameworks: Fcooper~\cite{fcooper}, V2X-ViT~\cite{v2x-vit}, Where2Comm~\cite{where2comm}, ERMVP~\cite{ermvp}, and DSRC~\cite{dsrc}.

Our method achieves state-of-the-art performance under both clean and corrupted conditions. On clean data, it outperforms the second-best method (DSRC) by +0.21/+2.58 on OPV2V (AP@0.5/AP@0.7) and +3.06/+2.30 on DAIR-V2X, demonstrating its superiority in ideal settings. More importantly, under corrupted conditions, our approach consistently surpasses the non-collaborative baseline by an average of +21.80/+38.64 on OPV2V and +11.17/+8.77 on DAIR-V2X across all seven corruption types, highlighting its robustness and the effectiveness of collaborative perception in adverse environments.

\noindent
\textbf{Sensitivity to Corruptions.}
\begin{figure}[!t]
    \centering
    \includegraphics[width=0.5\textwidth, trim=0cm 0cm 0cm 0cm, clip]{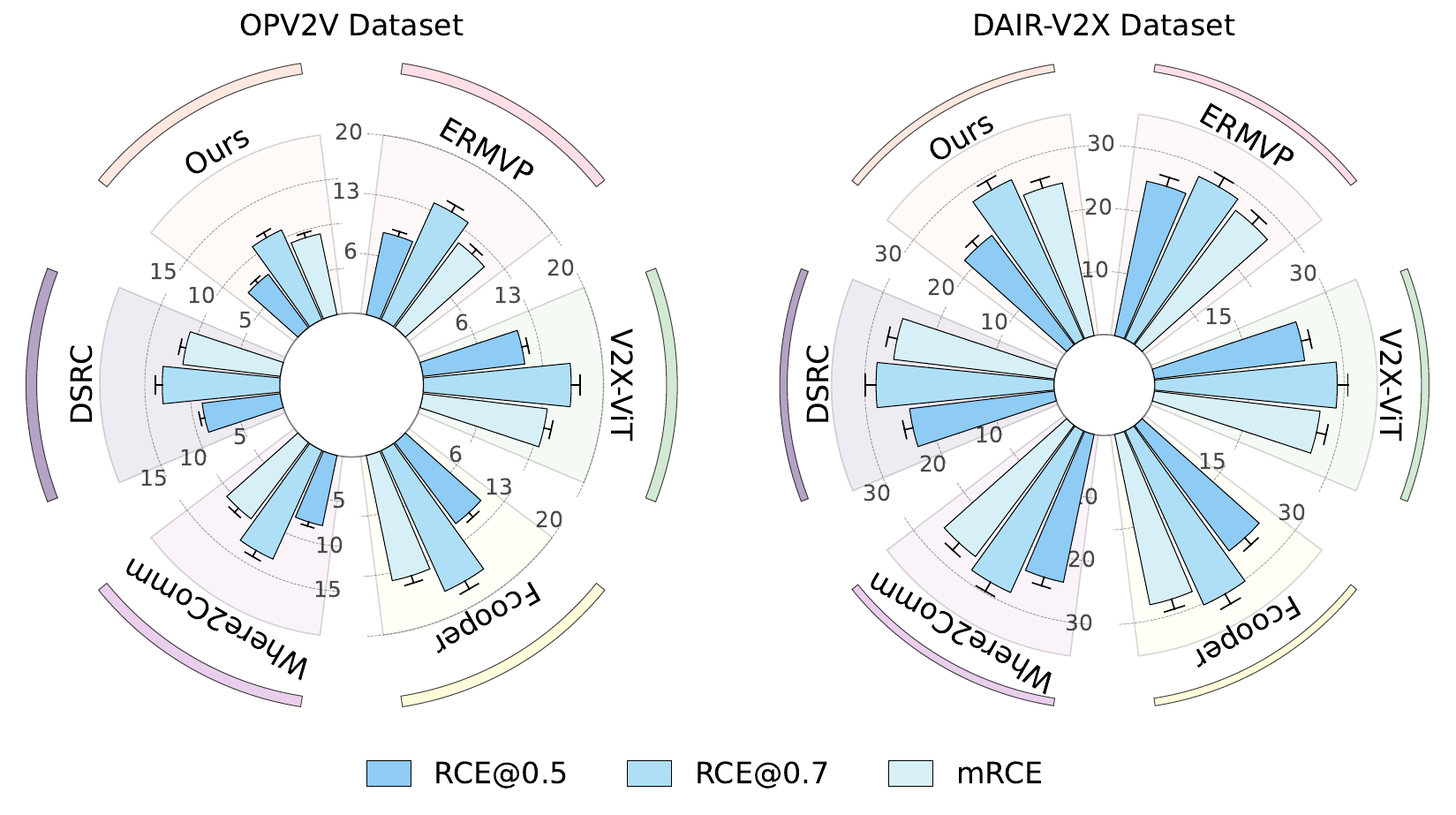}
    \caption{mRCE comparison across methods on OPV2V and DAIR-V2X.}
    \label{fig:flower}
\end{figure}
To evaluate robustness under real-world corruptions, we adopt the mean Relative Calibration Error (mRCE) as the primary metric, which is the arithmetic mean of RCE@0.5 and RCE@0.7. Specifically, RCE@0.5 and RCE@0.7 denote the average relative performance drop across all corruption types under AP@0.5 and AP@0.7, respectively. A lower mRCE indicates that the model better preserves its detection performance under input corruptions, i.e., higher robustness.

As shown in Fig.~\ref{fig:flower}, our method achieves the lowest mRCE on both benchmarks: 9.17\% on OPV2V, compared to 10.60\% for the second-best method (Where2Comm), and 24.69\% on DAIR-V2X, compared to 25.63\% for the runner-up (DSRC). These results demonstrate that our approach maintains superior performance stability across varying detection strictness and effectively mitigates degradation under diverse corruptions.

\noindent
\textbf{Sensitivity to Pose Noise.}
\begin{figure}[t]
    \centering
    \includegraphics[width=0.5\textwidth, trim=0cm 0.25cm 0cm 0.1cm, clip]{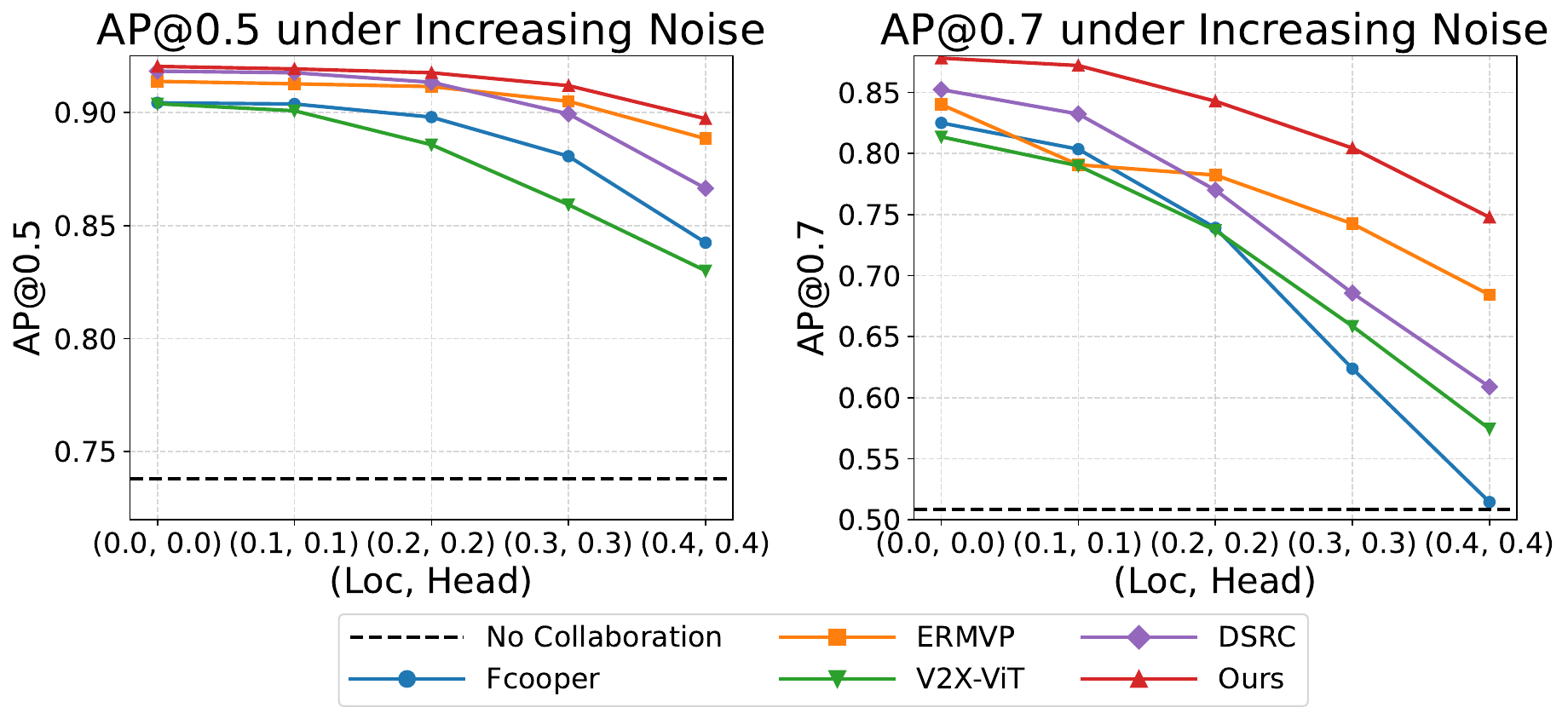}
    \caption{Detection performance under increasing pose noise on OPV2V. Loc and Head denote the standard deviations of localization error (in meters) and heading error (in radians), respectively.}
    \label{fig:robust}
\end{figure}
In real-world collaborative perception, relative poses between vehicles are often corrupted by estimation errors in localization and heading. These errors misalign spatially fused features, severely degrading collaborative detection performance. To evaluate robustness against this challenge, we inject Gaussian noise with increasing standard deviations into the localization and heading of all collaborating vehicles on OPV2V, simulating realistic pose uncertainties.

As shown in Fig.~\ref{fig:robust}, all methods degrade as pose noise increases from $(0,0)$ to $(0.4,0.4)$, confirming the adverse impact of pose uncertainty. However, our method consistently maintains the highest detection accuracy across all noise levels and exhibits the most gradual performance decay. Notably, at the highest noise level $(0.4,0.4)$, our method achieves an AP@0.7 of 74.77\%, outperforming the second-best method, ERMVP (68.42\%). In contrast, Fcooper’s performance drops sharply under high noise and approaches that of the No Collaboration baseline. This demonstrates that our framework is less sensitive to pose noise, maintaining high detection accuracy even under severe pose perturbations.

\subsection{Ablation Study}
\begin{table}[!t]
\centering
\caption{Ablation study of our proposed modules in teacher model and student model on DAIR-V2X.}
\label{tab:ablation}
\begin{tabular}{C C | C C}
\toprule
\multicolumn{4}{l}{\textit{Teacher}} \\
\midrule
\textbf{LGM} & \textbf{} & \textbf{AP@0.5} & \textbf{AP@0.7} \\
\midrule
 &  & 0.7631 & 0.6159 \\
$\checkmark$ & & 0.7750 & 0.6390 \\
\midrule
\multicolumn{4}{l}{\textit{Student (ours)}} \\
\midrule
\textbf{PKD} & \textbf{AGF} & \textbf{AP@0.5} & \textbf{AP@0.7} \\
\midrule
 & & 0.7375 & 0.5815 \\
\checkmark & & 0.7489 & 0.6038 \\
 & \checkmark & 0.7820 & 0.6265 \\
\checkmark & \checkmark & 0.7827 & 0.6392 \\
\bottomrule
\end{tabular}
\end{table}
To systematically evaluate the contribution of each component in our framework, we conduct ablation studies on the DAIR-V2X validation set.  
As shown in the Teacher section of Table~\ref{tab:ablation}, integrating LGM into the early-fusion teacher improves feature representation, boosting AP@0.7 from 0.6159 to 0.6390 (+2.31\%). Our student model is trained under the supervision of this enhanced teacher.
For the student model, we adopt DiscoNet~\cite{disconet} as the baseline and progressively incorporate our proposed modules to assess their individual and joint effects. Specifically: (1) PKD enables the student to learn rich and globally consistent semantic representations; (2) AGF performs fine-grained modulation during feature fusion, substantially improving robustness against noise and uncertainty.  

The full model equipped with both PKD and AGF achieves AP@0.5 = 78.27\% and AP@0.7 = 63.92\%, the highest performance across all settings, demonstrating the effectiveness and strong synergistic effect of the two modules. Notably, the student even outperforms the teacher, as the teacher relies on a static global view, while the student gains dynamic adaptation capability at inference.

\subsection{Qualitative Visualization}
\begin{figure}[t]
    \centering
    \includegraphics[width=0.55\textwidth, trim=2.2cm 2cm 1cm 2cm, clip]{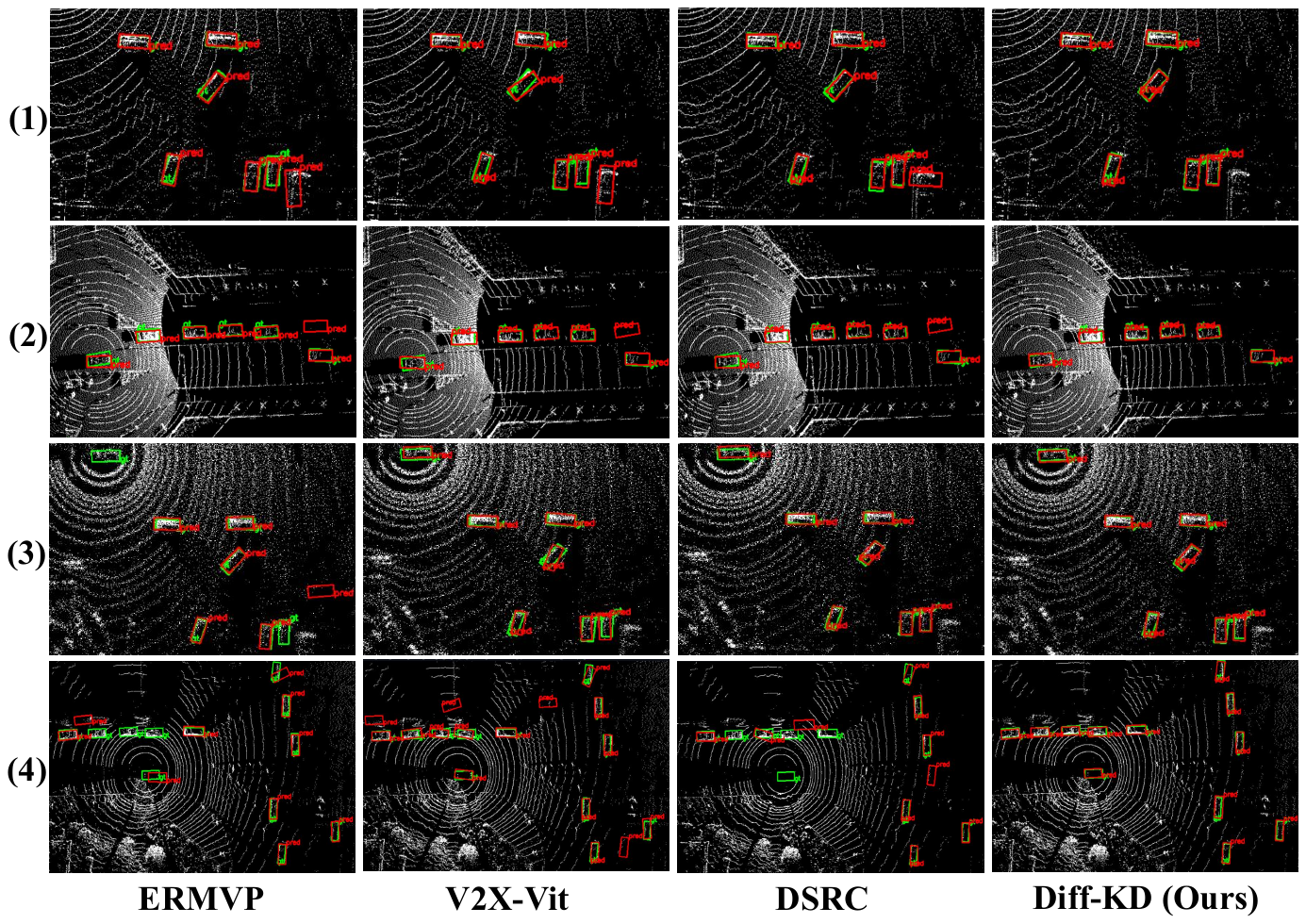}
    \caption{Qualitative comparison of detection results under four realistic sensor corruption scenarios: (1) clean condition, (2) echo, (3) motion blur, and (4) cross talk. Green boxes denote ground-truth annotations, and red boxes indicate model predictions.}
    \label{fig:vis}
\end{figure}
To further validate the robustness and generalization capability of our framework under real-world sensor corruption conditions, we present qualitative visualization results in Fig.~\ref{fig:vis}. Compared with other methods, our approach generates predictions that are not only more spatially accurate but also more complete in object coverage across all tested conditions. This qualitative evidence aligns consistently with our quantitative experiments.

\section{Conclusion}
We present Diff-KD, a diffusion-guided teacher–student framework that shifts collaborative perception from passive alignment of static, corrupted inputs to active, generative feature restoration. By integrating PKD for dynamic semantic recovery and AGF for uncertainty-aware collaboration, Diff-KD establishes a generative distillation paradigm that enables robust generalization under diverse real-world corruptions. Trained only on clean data, it consistently outperforms state-of-the-art methods across all corruption types on OPV2V and DAIR-V2X, paving a new pathway toward reliable and resilient multi-agent perception.

\bibliographystyle{IEEEtran}
\bibliography{main}

\end{document}